\documentclass[10pt,twocolumn,letterpaper]{article}

\usepackage{wacv}
\usepackage{times}
\usepackage{epsfig}
\usepackage{graphicx}
\usepackage{amsmath}
\usepackage{amssymb}
\usepackage{booktabs}

\usepackage[accsupp]{axessibility}
\usepackage{authblk}

\usepackage{cite}
\usepackage{graphicx}
\usepackage{cancel}
\usepackage{xfrac}
\usepackage{multirow}
\usepackage{pifont}
\usepackage{arydshln}
\usepackage{adjustbox}
\usepackage{subcaption}
\usepackage{wrapfig}
\usepackage{makecell}
\usepackage{arydshln}
\usepackage{xcolor}

\usepackage{color}
\usepackage[ruled,vlined]{algorithm2e}
\newcommand{\PyComment}[1]{\ttfamily\textcolor{commentcolor}{\# #1}}  
\newcommand{\PyCode}[1]{\ttfamily\textcolor{black}{#1}} 
\definecolor{commentcolor}{RGB}{30,100,100}   

\makeatletter
\def\adl@drawiv#1#2#3{%
        \hskip.5\tabcolsep
        \xleaders#3{#2.5\@tempdimb #1{1}#2.5\@tempdimb}%
                #2\z@ plus1fil minus1fil\relax
        \hskip.5\tabcolsep}
\newcommand{\cdashlinelr}[1]{%
  \noalign{\vskip\aboverulesep
           \global\let\@dashdrawstore\adl@draw
           \global\let\adl@draw\adl@drawiv}
  \cdashline{#1}
  \noalign{\global\let\adl@draw\@dashdrawstore
           \vskip\belowrulesep}}
\makeatother

\newsavebox\CBox
\def\textBF#1{\sbox\CBox{#1}\resizebox{\wd\CBox}{\ht\CBox}{\textbf{#1}}}

\newcommand{\figref}[1]{Figure~\ref{#1}}
\newcommand{\tabref}[1]{Table~\ref{#1}}
\newcommand{\algoref}[1]{Algorithm~\ref{#1}}

\definecolor{lg}{RGB}{165,0,52}

\newcommand{\ra}[1]{\renewcommand{\arraystretch}{#1}}

\usepackage{tabulary}
\newcolumntype{x}[1]{>{\centering\arraybackslash}p{#1pt}}

\newlength\savewidth

\newcommand\blfootnote[1]{%
  \begingroup
  \renewcommand\thefootnote{}\footnote{#1}%
  \addtocounter{footnote}{-1}%
  \endgroup
}

\def\wacvPaperID{1988} 

\wacvalgorithmstrack   

\wacvfinalcopy 

\ifwacvfinal
\usepackage[breaklinks=true,bookmarks=false,colorlinks,breaklinks=true,citecolor=lg]{hyperref}
\else
\usepackage[pagebackref=true,breaklinks=true,colorlinks,bookmarks=false]{hyperref}
\fi

\begin{document}

\title{Enriched CNN-Transformer Feature Aggregation Networks for Super-Resolution}

\author{Jinsu Yoo$^{1*}$~~~Taehoon Kim$^2$~~~Sihaeng Lee$^2$~~~Seung Hwan Kim$^2$~~~Honglak Lee$^2$~~~Tae Hyun Kim$^{1}$\\
$^1$Hanyang University~~~$^2$LG AI Research
}

\maketitle
\thispagestyle{empty}

\begin{abstract}
Recent transformer-based super-resolution (SR) methods have achieved promising results against conventional CNN-based methods.
However, these approaches suffer from essential shortsightedness created by only utilizing the standard self-attention-based reasoning.
In this paper, we introduce an effective hybrid SR network to aggregate enriched features, including local features from CNNs and long-range multi-scale dependencies captured by transformers.
Specifically, our network comprises transformer and convolutional branches, which synergetically complement each representation during the restoration procedure.
Furthermore, we propose a cross-scale token attention module, allowing the transformer branch to exploit the informative relationships among tokens across different scales efficiently.
Our proposed method achieves state-of-the-art SR results on numerous benchmark datasets.
\end{abstract}

\blfootnote{$^*$Work done while interning at LG AI Research. Code is available at: \href{https://github.com/jinsuyoo/act}{https://github.com/jinsuyoo/act}.}

\section{Introduction}

Super-resolution (SR) is a longstanding problem that aims to restore a high-resolution (HR) image from the given low-resolution (LR) image.
With the advancements in deep learning, various CNN-based SR methods are introduced~\cite{dong2015srcnn, kim2016vdsr, ledig2017srgan, lim2017edsr, zhang2018rcan, niu2020han, mei2021nlsa}. 
The emergence of CNN-based SR networks have verified the efficiency in processing 2D images with the inductive bias (\ie, local connectivity and translation invariance).
However, such architectures have certain limitations in exploiting global information~\cite{dosovitskiy2021vit} or restoring weak texture details~\cite{chen2021ipt}.

To solve the problem, SR researchers~\cite{chen2021ipt, liang2021swinir, cao2021vsrt, liang2022vrt} have recently applied Vision Transfomer (ViT)~\cite{dosovitskiy2021vit} to SR architectures.
A self-attention mechanism, the core component of ViT, enables the network to capture long-range spatial dependencies within an image.
In particular, ViT inherently contains superiority over CNN in exploiting self-similar patches within the input image by calculating similarity among tokens (patches) over the entire image region.
Built upon standard~\cite{dosovitskiy2021vit} or sliding window-based~\cite{liu2021swin} self-attention, such transformer-based SR networks have remarkably improved the restoration performance.

However, existing approaches~\cite{chen2021ipt, liang2021swinir, cao2021vsrt, liang2022vrt} suffer from features limitedly extracted through a certain type of self-attention mechanism.
Accordingly, the networks cannot utilize various sets of features, such as local or multi-scale features, which are proven effective for SR~\cite{mei2020csnln, guo2020drn}.
This problem raises three concerns in building transformer-based SR architecture.
First, although ViTs extract non-local dependencies better, CNN is still a preferable way to efficiently leverage repeated local information within an image~\cite{wu2021cvt, d2021convit, chen2021mobileformer}.
Next, restoring images solely with tokenized image patches can cause undesired artifacts at the token boundaries.
While tokenization with large overlapping alleviates such a problem, this approach will considerably increase the computational cost of self-attention.
Finally, tokenization with an identical token size limits the exploitation of multi-scale relationships among tokens.
Notably, reasoning across different scaled patches (tokens) is exceptionally beneficial in SR~\cite{shocher2018zssr, mei2020csnln} as it utilizes the internal self-similar patches~\cite{zontak2011internal} within an image.

In this paper, we propose to \textbf{A}ggregate enriched features extracted from both \textbf{C}NN and \textbf{T}ransformer (\textbf{ACT}) mechanisms and introduce an effective hybrid architecture that takes advantage of multi-scale local and non-local information.
Specifically, we construct two different branches (\ie, CNN and transformer branches) and fuse the intermediate representations during the SR procedure.
Consequently, local features extracted from the CNN branch and long-range dependencies captured in the transformer branch are progressively fused to complement each other and extract robust features.
Furthermore, we propose a \textbf{C}ross-\textbf{S}cale \textbf{T}oken \textbf{A}ttention module (\textbf{CSTA}) inside the transformer branch, which overcomes the limitation of prior transformer-based SR methods~\cite{chen2021ipt, liang2021swinir, cao2021vsrt} in exploiting multi-scale features.
Inspired by re-tokenization~\cite{yuan2021t2t}, CSTA efficiently generates multi-scale tokens and enables the network to learn multi-scale relationships.
Lastly, we investigate the necessity of commonly used techniques with transformers, such as positional embeddings, for SR task.

The proposed network, ACT, achieves state-of-the-art SR performances on several benchmark datasets and even outperforms recent transformer-based SR methods~\cite{chen2021ipt, liang2021swinir}.
Notably, our ACT extraordinarily improves SR quality when test image contains numerous repeating patches.
To sum up, our contributions are presented as follows:
\begin{itemize}
    \item We introduce a novel hybridized SR method, combining CNN and ViT, to effectively aggregate an enriched set of local and non-local features.
    \item We propose a cross-scale token attention module to leverage multi-scale token representations efficiently.
    \item Extensive experiments on numerous benchmark SR datasets demonstrate the superiority of our method.
\end{itemize}
    
\section{Related Work}

\paragraph{CNN-based SR:}

Several SR architectures have been designed upon convolutional layer to extract beneficial features from a given LR image~\cite{dong2015srcnn, kim2016vdsr,lim2017edsr,zhang2018rcan,niu2020han,dai2019san,guo2020drn,ledig2017srgan}.
In particular, researchers have focused on exploiting a non-local internal information (\ie, patch-recurrence~\cite{zontak2011internal}) within an image~\cite{liu2018nlrn, zhang2019rnan, mei2020csnln, zhou2020ignn, mei2021nlsa}.
This is achieved by adding various modules such as recurrent-based module~\cite{liu2018nlrn, mei2020csnln} or graph neural networks~\cite{zhou2020ignn}.
Constructed on top of baseline architectures (\eg, EDSR~\cite{lim2017edsr} or RCAN~\cite{zhang2018rcan}), these methods improve the SR performance further.
Our work is inspired by the recent success of ViT~\cite{dosovitskiy2021vit} in exploiting global representations within an image.
Unlike previous studies, we separate CNN and Transformer branches to leverage local and non-local features individually and fuse the intermediate representations to compensate for each other.

\paragraph{ViT-based SR:}

Recent ViT-based networks have escalated the performance for various computer vision tasks~\cite{carion2020detr, ranftl2021dpt, liu2021swin, wang2021uformer, liu2021swin, wu2021cvt, guo2021cmt, li2021localvit, d2021convit}.
Moreover, researchers have explored building the architecture for image restoration~\cite{wang2022uformer, chen2021ipt, liang2021swinir}.
For image SR, Chen~\etal~\cite{chen2021ipt} proposed a pure ViT-based network~\cite{dosovitskiy2021vit} to handle various image restoration tasks, including denoising, deraining, and SR.
First, the network is pre-trained with a multi-task learning scheme, including the entire tasks.
Then, the pre-trained network is fine-tuned to the desired task (\eg, $\times2$ SR).
Instead of the standard self-attention, Liang~\etal~\cite{liang2021swinir} adapted the Swin Transformer block \cite{liu2021swin} and included convolutional layers inside the block to impose the local connectivity.
Motivated to overcome the limitations within standard self-attention-based networks (\ie, IPT~\cite{chen2021ipt}), we model an effective hybridized architecture to aggregate enriched features across different scales. 

\paragraph{Multi-scale ViTs:}

Many prior arts have studied multi-scale token representations for transformer~\cite{chen2021crossvit, wang2021pvt, wang2022uformer}.
Among them, Chen \etal \cite{chen2021crossvit} explicitly tokenized an image with different token sizes, while Wang \etal \cite{wang2021pvt} constructed the pyramid architecture.
Differently, we basically maintain a single token representation.
Then, we utilize channel-wise splitting and re-tokenization~\cite{yuan2021t2t} to efficiently generate multi-scale tokens and reason across them.
Furthermore, our cross-scale attention aims to exploit self-similar patches across scales~\cite{zontak2011internal} within an input image.

\paragraph{Multi-branch architectures:}

Numerous works have constructed multi-branch networks~\cite{simonyan2014two, feichtenhofer2016stresnet, feichtenhofer2017spatiotemporal, feichtenhofer2019slowfast, chen2021mobileformer} to handle several input data containing different information effectively.
For SR, Jin~\etal~\cite{jin2019procar} and Isobe~\etal~\cite{isobe2020rsdn} decomposed the input image/video into structural and texture information to advantageously restore the missing high-frequency details. 
In this study, we borrow the recent ViT to enhance global connectivity and model capacity.

\section{Proposed Method}

\begin{figure*}[t]
    \centering
    \includegraphics[width=0.9\linewidth]{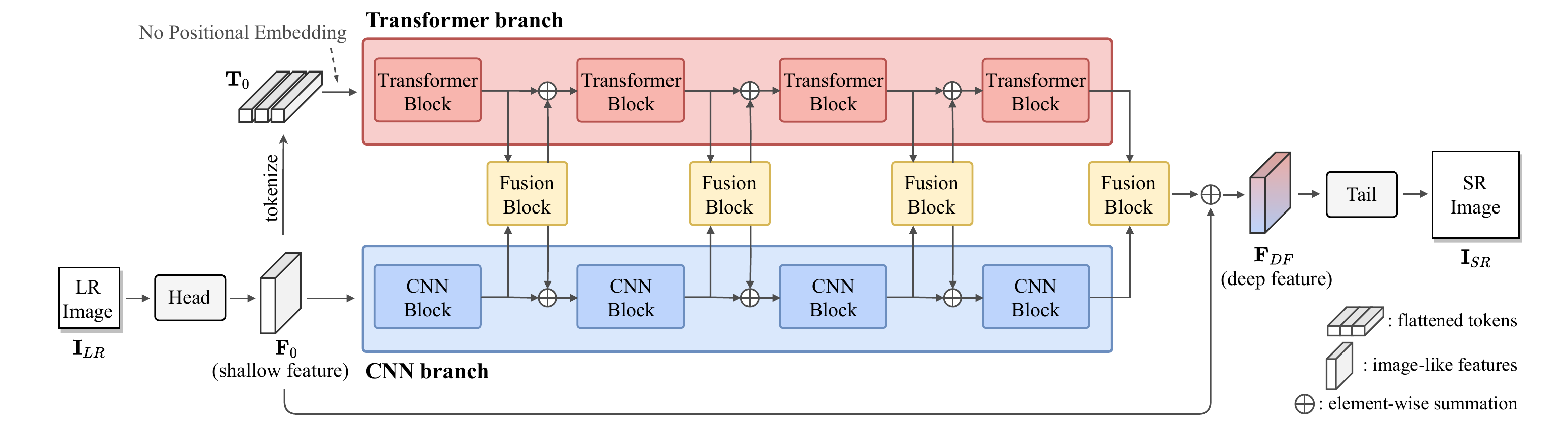}
    \caption{The overall flow of ACT. The input image goes through two separate branches constructed with a CNN and ViT. Each branch extracts local features and global representations and actively exchanges beneficial information during the intermediate fusion. The final SR result is acquired by aggregating enriched representations.}
    \label{fig:overall}
\end{figure*}

In this section, we present ACT, which leverages both CNN and transformer, in detail.
ACT is composed of a shallow feature extraction module (head), transformer/CNN modules (body), and a high-quality image reconstruction module (tail).
\figref{fig:overall} illustrates the overall structure. 

\subsection{Head}
\label{head}

First, head module $H_{Head}$ extracts shallow feature $\mathbf{F}_{0}$ from a given low-resolution input image $\mathbf{I}_{LR}$ as:
\begin{equation}
    \mathbf{F}_{0} = H_{Head}(\mathbf{I}_{LR}),
\end{equation}
where $H_{Head}$ is composed of two residual convolution blocks as suggested in previous works~\cite{lim2017edsr, zhang2018rcan, chen2021ipt}.

\subsection{Body}
\label{body}

Next, the extracted feature $\mathbf{F}_{0}$ is passed to the body module to acquire robust features as:
\begin{equation}
    \mathbf{F}_{DF} = H_{Body}(\mathbf{F}_{0}) + \mathbf{F}_{0},
\end{equation}
where $\mathbf{F}_{DF}$ indicates deep feature, and $H_{Body}$ contains proposed two-stream branches to extract residual features correspondingly.

\subsubsection{CNN branch}
\label{cb}
In the body module, convolutional kernels in the CNN branch slide over the image-like features with a stride of 1.
Such a procedure compensates for the transformer branch's lack of intrinsic inductive bias.
In building our CNN branch, we adopt the residual channel attention module (RCAB)~\cite{zhang2018rcan}, which has been widely used in the recent SR approaches~\cite{niu2020han, guo2020drn}.
Specifically, we stack $N$ CNN Blocks, which yields the following:
\begin{equation}
    \mathbf{F}_{i} = H_{CB}^{i} (\mathbf{F}_{i-1}), \quad 1 \leq i \leq N,
\end{equation}
where $H_{CB}^i$ denotes CNN Block including RCAB modules, and $\mathbf{F}_{i},$ represents extracted CNN feature at $i$th CNN Block.

\subsubsection{Transformer branch}
\label{tb}

\begin{figure*}[t]
    \centering
    \includegraphics[width=1\linewidth]{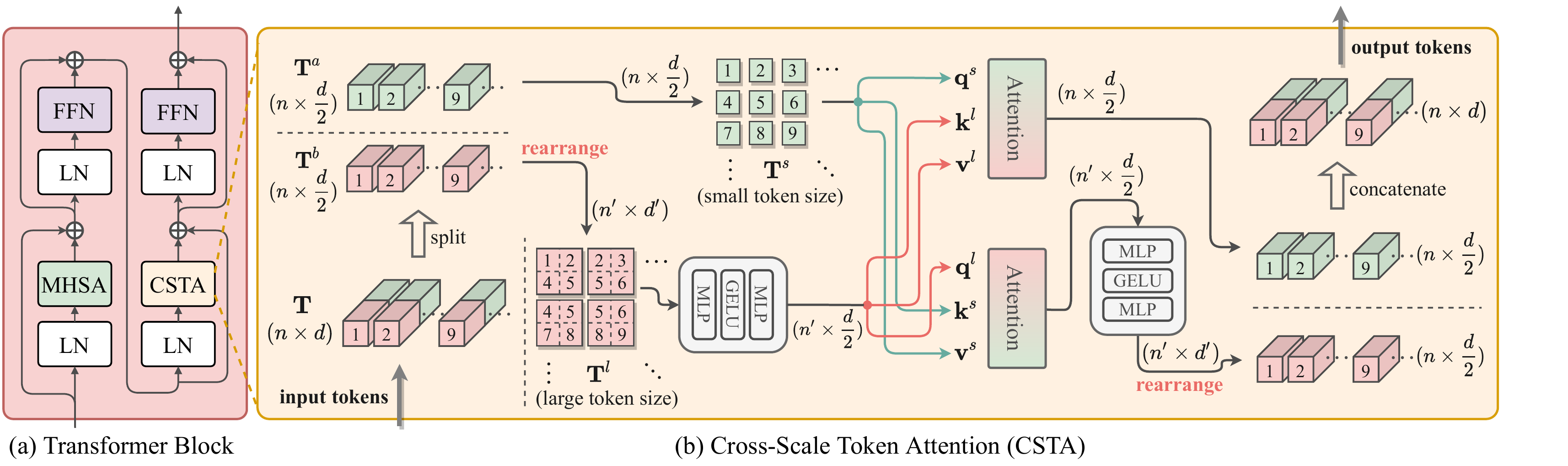}
    \caption{(a) Our Transformer Block includes multi-head self-attention ($\text{MHSA}$) and cross-scale token attention ($\text{CSTA}$). (b) $\text{CSTA}$ effectively exploits information across different scaled tokens by channel-wise splitting and token rearrangement. Two different token embeddings exchange keys and values for the attention operation.}
    \label{fig:block}
\end{figure*}

We construct transformer branch in body module based on standard multi-head self-attention ($\text{MHSA}$) \cite{dosovitskiy2021vit, chen2021ipt}. 
Moreover, we propose to add cross-scale token attention ($\text{CSTA}$) modules in the transformer branch to exploit repeating structures within an input image across different scales.

First, we tokenize image-like shallow feature $\mathbf{F}_0 \in \mathbb{R}^{c \times h \times w}$ into non-overlapping $n$ tokens $\mathbf{T}_{0} \in \mathbb{R}^{n \times d}$, where $d$ is dimension of each token vector.
Notably, $n = \sfrac{h}{t} \times \sfrac{w}{t}$ and $d = c \cdot t^2$, where $t$ is the token size.
Moreover, unlike previous ViTs \cite{dosovitskiy2021vit, chen2021ipt}, we observe that positional information becomes insignificant in SR. Thus, we do not add positional embeddings to tokens. 

Then, acquired tokens are fed into the transformer branch, including $N$ Transformer Blocks, which are symmetric to the CNN branch as:
\begin{equation}
    \mathbf{T}_{i} = H_{TB}^{i} (\mathbf{T}_{i-1}), \quad 1 \leq i \leq N,
\end{equation}
where $\mathbf{T}_{i}$ indicates extracted token representations at $i$th Transformer Block $H_{TB}^i$.
As depicted in \figref{fig:block}a, each Transformer Block includes two sequential attention operations: multi-head self-attention (MHSA) \cite{dosovitskiy2021vit, chen2021ipt} and cross-scale token attention (CSTA), which yield:
\begin{equation}
    \mathbf{T}_{i} = \text{FFN}(\text{CSTA}(\mathbf{T}^{\prime}_{i-1})), \mathbf{T}^{\prime}_{i-1} = \text{FFN}(\text{MHSA}(\mathbf{T}_{i-1})),
\end{equation}
where $\text{FFN}$ (feed forward network) includes two MLP layers with expansion ratio $r$ with GELU activation function \cite{hendrycks2016gelu} in middle of the layers.
Here, we omit layer normalization (LN)~\cite{ba2016ln} and skip connections for brevity, and details of our $\text{CSTA}$ are as follows.

\paragraph{Cross-scale token attention (CSTA):}

Standard $\text{MHSA}$ \cite{vaswani2017attention, dosovitskiy2021vit} performs attention operation by projecting queries, keys, and values from the same source of single-scale tokens.
In addition to self-attention, we propose to exploit information from tokens across different scales,
and we illustrate operational flow of proposed $\text{CSTA}$ module in \figref{fig:block}b.
Concretely, we split input token embeddings $\mathbf{T} \in \mathbb{R}^{n \times d}$ of CSTA along the last (\ie, channel) axis into two,
and we represent them as $\textbf{T}^a \in \mathbb{R}^{n \times \sfrac{d}{2}}$ and $\textbf{T}^b \in \mathbb{R}^{n \times \sfrac{d}{2}}$.
Then, we generate $\mathbf{T}^{s} \in \mathbb{R}^{n \times \sfrac{d}{2}}$ and $\mathbf{T}^{l} \in \mathbb{R}^{n^\prime \times d^\prime}$, which include $n$ tokens from $\textbf{T}^a$ and $n^{\prime}$ tokens by rearranging $\textbf{T}^b$, respectively.
In practice, we use $\mathbf{T}^{a}$ for $\mathbf{T}^{s}$ as is, and re-tokenize \cite{yuan2021t2t} $\mathbf{T}^{b}$ to generate $\mathbf{T}^{l}$ with larger token size and overlapping.
Here, we can control number of tokens in $\mathbf{T}^{l}$ by setting as $n^{\prime} = \left\lfloor \frac{h - t^{\prime}}{s^{\prime}} + 1 \right\rfloor \times \left\lfloor \frac{w - t^{\prime}}{s^{\prime}} + 1 \right\rfloor$, where $s^{\prime}$ is stride and $t^{\prime}$ denotes token size, and token dimension is $d^{\prime} = (\sfrac{c}{2})\cdot {t^{\prime}}^2 = \sfrac{(d\cdot {t^{\prime}}^2)}{2t^2}$.
One can acquire numerous tokens of large size by overlapping, which enables the network to enjoy patch-recurrence across scales actively.
Notably, large tokens are essential for reasoning repeating patches across scales during the CSTA procedure.

In particular, to effectively exploit self-similar patches across different scales and pass a larger patch's information to small but self-similar ones,
we produce a smaller number of tokens  (\ie, $n^{\prime} < n$) with a relatively larger token size (\ie, $t^{\prime} > t$), and then compute cross-scale attention scores between tokens in both $\mathbf{T}^{s}$ and $\mathbf{T}^{l}$.

Specifically, we generate queries, keys and values in $\mathbf{T}^{s}$ and $\mathbf{T}^{l}$: ($\mathbf{q}^{s}\in\mathbb{R}^{n \times \sfrac{d}{2}}$, $\mathbf{k}^{s}\in\mathbb{R}^{n \times \sfrac{d}{2}}, \mathbf{v}^{s}\in\mathbb{R}^{n \times \sfrac{d}{2}}$) from $\mathbf{T}^{s}$, and ($\mathbf{q}^{l}\in\mathbb{R}^{n' \times \sfrac{d}{2}}$, $\mathbf{k}^{l}\in\mathbb{R}^{n' \times \sfrac{d}{2}}, \mathbf{v}^{l}\in\mathbb{R}^{n' \times \sfrac{d}{2}}$) from $\mathbf{T}^{l}$.
Next, we carry out attention operation~\cite{vaswani2017attention} using triplets ($\mathbf{q}^{l}, \mathbf{k}^{s}, \mathbf{v}^{s}$) and ($\mathbf{q}^{s}, \mathbf{k}^{l}, \mathbf{v}^{l}$) as inputs by exchanging key-value pairs from each other.
Notably, last dimension of queries, keys, and values from $\mathbf{T}^{l}$ is lessened from $d^{\prime}$ to $\sfrac{d}{2}$ by projections for the attention operation, 
and attention result for $\mathbf{T}^{l}$ is re-projected to dimension of $n^{\prime} \times d^{\prime}$, 
then rearranged to dimension of $n \times \frac{d}{2}$.
Finally, we generate an output token representation of CSTA by concatenating attention results.

Our $\text{CSTA}$ can exploit cross-scale information without additional high overhead.
More specifically,
computation costs of $\text{MHSA}$ and $\text{CSTA}$ are as follows:
\begin{equation}
\begin{split}
&\mathcal{O}(MHSA) = n^{2} \cdot d + n \cdot d^{2}\\
&\mathcal{O}(CSTA) = (n \cdot n^{\prime}) \cdot d + (n + n^{\prime}) \cdot d^{2},
\end{split}
\end{equation}
and computational cost of $\text{CSTA}$ is competitive with $\text{MHSA}$ because $n > n^{\prime}$.

The notion of our $\text{CSTA}$ module is computing attention scores across scales in a high-dimensional feature space and explicit reasoning across multi-scale tokens.
Thus, our network can utilize recurring patch information across different scales within the input image~\cite{shocher2018zssr, mei2020csnln}, while conventional $\text{MHSA}$ limitedly extract informative cross-scale cues.

\subsubsection{Multi-branch feature aggregation}
\label{fb}

\begin{figure*}[t]
    \centering
    \includegraphics[width=0.8\linewidth]{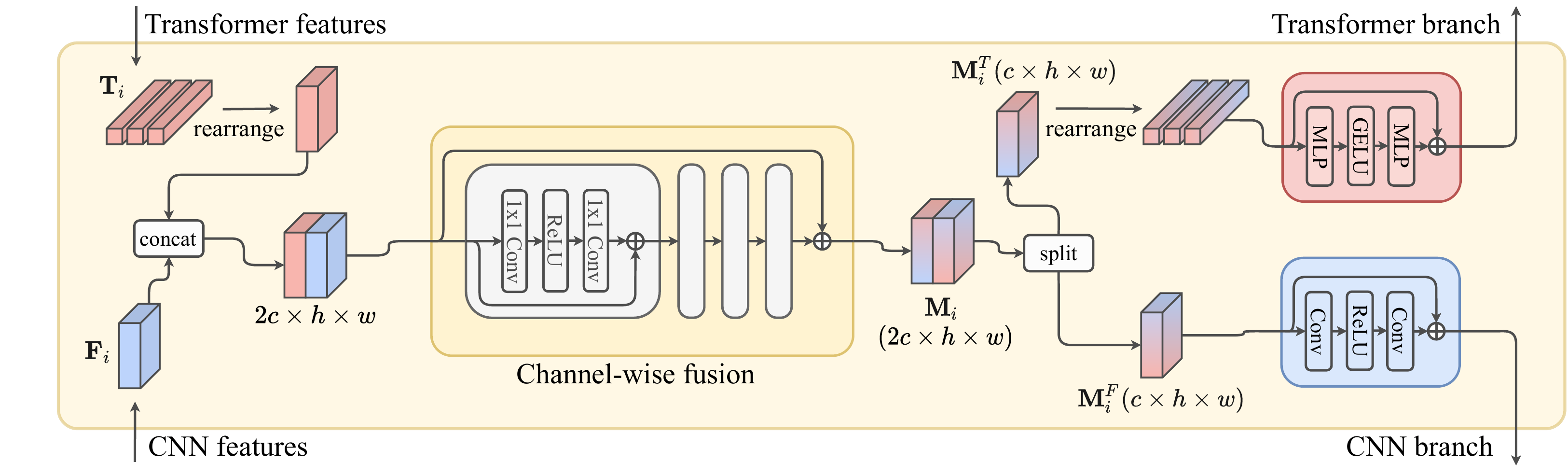}
    \caption{Our Fusion Block. Concatenated features from two branches are fused with 1$\times$1 convolutions. Then, complemented information is bidirectionally transferred to the original branches.}
    \label{fig:fb}
\end{figure*}

We bidirectionally connect intermediate features extracted from independent branches.
\figref{fig:fb} depicts our Fusion Block.
Concretely, given intermediate features $\mathbf{T}_i$ and $\mathbf{F}_i$ from $i$th CNN Block and Transformer Block, we aggregate feature maps by using Fusion Block $H_{fuse}$ as:
\begin{equation}
\label{eq:7}
    \mathbf{M}_{i} = H_{fuse}^{i} (\mathtt{rearrange}(\mathbf{T}_{i}) \mathbin\Vert \mathbf{F}_{i}), \quad 1 \leq i \leq N,
\end{equation}
where $\mathbf{M}_{i} \in \mathbb{R}^{2c \times h \times w}$ denotes fused features, and $\mathtt{rearrange}$ and $\mathbin\Vert$ represent image-like rearrangement and concatenation, respectively.
We build our Fusion Block $H_{fuse}$ with $1 \times 1$ convolutional blocks for a channel-wise fusion.
Except for last Fusion Block (\ie, $i=N$), fused features $\mathbf{M}_{i}$ are split into two features along channel dimension, \ie, $\mathbf{M}^{T}_{i} \in \mathbb{R}^{c \times h \times w}$ and $\mathbf{M}^{F}_{i} \in \mathbb{R}^{c \times h \times w}$, followed by MLP blocks and convolutional blocks, respectively.
Then, each fused feature flows back to each branch and is individually added to the original input feature $\mathbf{T}_i$ and $\mathbf{F}_i$.
Fused feature $\mathbf{M}_{N}$ at last Fusion Block takes a single $3\times3$ convolution layer to resize channel dimension from $2c$ to $c$.
The extracted deep residual feature is added to $\mathbf{F}_{0}$ and produces a deep feature $\textbf{F}_{DF}$.
Subsequently, $\textbf{F}_{DF}$ is transferred to the final tail module.

\subsection{Tail}
\label{tail}

For the last step, aggregated feature $\mathbf{F}_{DF}$ is upscaled and reconstructed through tail module $H_{Tail}$ and produces the final SR result as:
\begin{equation}
    \mathbf{I}_{SR} = H_{Tail}(\mathbf{F}_{DF}).
\end{equation}
$H_{Tail}$ includes PixelShuffle \cite{shi2016pixelshuffle} operation, which upscales feature maps by rearranging channel-wise features to the spatial dimension, followed by a single convolution layer to predict the final SR result.

\section{Experiments}

In this section, we quantitatively and qualitatively demonstrate the superiority of ACT.

\subsection{Implementation details}

\paragraph{Datasets and evaluation metrics:}

Following previous works~\cite{dong2015srcnn, chen2021ipt}, we train our network with the ImageNet dataset~\cite{deng2009imagenet}.
Therefore, the transformer can fully utilize its representation capability~\cite{chen2021ipt}.
Specifically, we use approximately 1.3M images such that the length of the shortest axis exceeds 400 pixels as ground-truth HR images. 
Input LR images are generated by downscaling HR images using bicubic interpolation.
Moreover, we evaluate performance of SR networks on conventional SR benchmark datasets: Set5 \cite{bevilacqua2012set5}, Set14 \cite{zeyde2010set14}, B100 \cite{martin2001b100}, Urban100 \cite{huang2015selfex}, and Manga109 \cite{matsui2017manga109}.
The experimental results are evaluated with two metrics, namely peak signal-to-noise ratio (PSNR) and structural similarity (SSIM), on Y channel in YCbCr color space following baselines \cite{lim2017edsr, zhang2018rcan, chen2021ipt}.

\vspace{-0.2cm}
\paragraph{Hyperparameters:}

\begin{table}[t]
\centering
\ra{1.0}
\scriptsize
\begin{adjustbox}{width=0.8\linewidth}
\begin{tabular}{l|x{20}x{20}|x{20}x{20}}
\toprule
 & \multicolumn{2}{c|}{Standard ViT~\cite{dosovitskiy2021vit}} & \multicolumn{2}{c}{ACT (Ours)} \\ \midrule
Learnable PE~\cite{chen2021ipt} & w/ & w/o & w/ & w/o \\ \midrule
Set5~\cite{bevilacqua2012set5} & 38.31 & \textBF{38.33} & 38.43 & \textBF{38.46} \\
Set14~\cite{zeyde2010set14} & 34.29 & \textBF{34.33} & 34.57 & \textBF{34.60} \\ \bottomrule
\end{tabular}
\end{adjustbox}
\caption{Ablation on the impact of positional embeddings for SR, reported in PSNR value. PE indicates positional embeddings.}
\label{tab:pe}
\end{table}

We have four CNN Blocks in the CNN branch and four Transformer Blocks in the transformer branch (\ie, $N=4$).
Each CNN Block includes 12 RCAB modules with channel size $c = 64$.
We use $d$ = 576, and expansion ratio $r$ inside $\text{FFN}$ module is set to 4.
For Fusion Block, we stack four $1 \times 1$ residual blocks~\cite{he2016resnet}.
During training, input LR patches are fixed to $48 \times 48$, and token size is $3 \times 3$ (\ie, $t=3$).
We also use conventional data augmentation techniques, such as rotation (${90}^{\circ}$, ${180}^{\circ}$, and ${270}^{\circ}$) and horizontal flipping.
Large token size $t^{\prime}$, stride $s^{\prime}$, and $d'$ for $\text{CSTA}$ module are set to 6, 3, and 1152, respectively.
Moreover, we use Adam optimizer \cite{kingma2014adam} to train our network
and minimize $L_1$ loss following previous studies \cite{zhang2018rcan, niu2020han, chen2021ipt}.
We train our network for 150 epochs with a batch size of 512. 
The initial learning rate is $10^{-4}$ ($\beta_1 = 0.9$ and $\beta_2 = 0.999$) and we reduce it by half every 50 epochs.
We implement our model using the PyTorch framework with eight NVIDIA A100 GPUs.
    
\subsection{Ablation studies}
\label{ablation}

\subsubsection{Impact of positional embeddings}
\label{sec:pe}

Unlike high-level vision tasks (\eg, classification), the transformer in SR utilizes a relatively small patch size. 
To investigate the necessity of positional embeddings for SR, we train two types of SR networks by using 1) MHSA instead of CSTA and removing the CNN branch (\ie, standard ViT \cite{dosovitskiy2021vit}) and 2) our ACT.
Networks are trained with and without learnable positional embeddings \cite{dosovitskiy2021vit, carion2020detr, chen2021ipt}, and results are provided in \tabref{tab:pe}.
We observe that transformer without positional embeddings does not degrade SR performance.
According to our observation, we do not use positional embeddings in the remainder of our experiments.
    
\subsubsection{Impact of fusing strategies}

\begin{table}[t]\centering
\ra{1.0}
\begin{adjustbox}{width=1\linewidth}
\footnotesize
\begin{tabular}{l|x{25}x{25}|x{25}x{25}x{25}x{25}}\toprule
 & \multicolumn{2}{c|}{\textBF{Single-stream}} & \multicolumn{4}{c}{\textBF{Two-stream}} \\ \midrule
Transformer & w/ & w/o & w/ & w/ & w/ & w/ \\
CNN & w/o & w/ & w/ & w/ & w/ & w/ \\
Fusion & w/o & w/o & w/o & T $\rightarrow$ C & C $\rightarrow$ T & T $\leftrightarrow$ C \\ \midrule
\# Params. & 32.8M & 4.3M & 36.6M & 37.4M & 45.1M & 45.3M \\ \midrule


Set14~\cite{zeyde2010set14} & 34.33 & 34.21 & 34.32 & 34.38 & 34.44 & \textBF{34.45} \\ 
Manga109~\cite{matsui2017manga109} & 39.48 & 39.46 & 39.56 & 39.71 & 39.77 & \textBF{39.85} \\ \bottomrule
\end{tabular}
\end{adjustbox}
\caption{Ablation experiments on various architectural choices \wrt PSNR metric. 
T and C means Transformer and CNN branches respectively. $\rightarrow$ indicates unidirectional flow from left to right, and $\leftrightarrow$ indicates bidirectional flow.} 
\label{tab:design}
\end{table}

\begin{table*}[t]
    \begin{subtable}[t]{0.32\textwidth}
    \centering
        \begin{adjustbox}{width=1\linewidth}
        \begin{tabular}[t]{lcc}\toprule
            \textBF{Method} & \textBF{\# Params.}& \textBF{Set14/Urban100} \\ \midrule
            MHSA only & 45.3M & 34.45/33.93\\ 
            MHSA + CSTA & 46.0M & \textBF{34.60}/\textBF{34.07} \\ 
            CSTA only & 46.7M & 34.51/33.92\\ \bottomrule
        \end{tabular}
        \end{adjustbox}
    \caption{MHSA vs. CSTA. Utilizing both attentions performs better than MHSA or CSTA alone.}
    \label{cstamhsa}
    \end{subtable}
    \hfill
    \begin{subtable}[t]{0.35\textwidth}
    \centering
        \begin{adjustbox}{width=1\linewidth}
        \begin{tabular}[t]{l|c|ccc}\toprule
            \textBF{CSTA} & w/o & \multicolumn{3}{c}{w/} \\ \midrule
            \textBF{Stride} ($s^{\prime}$) & - & \multicolumn{1}{c}{5} & \multicolumn{1}{c}{4} & 3 \\
            \textBF{\# Large tokens} ($n^{\prime}$) & - & \multicolumn{1}{c}{81} & \multicolumn{1}{c}{121} & 225 \\ \midrule
            \textBF{FLOPs} & 22.2G & \multicolumn{1}{c}{\textBF{21.4G}} & \multicolumn{1}{c}{21.6G} & 22.2G \\ 
            \textBF{PSNR on Set14} & 34.45 & \multicolumn{1}{c}{34.51} & \multicolumn{1}{c}{34.55} & \textBF{34.60} \\ \bottomrule
        \end{tabular}
        \end{adjustbox}
    \caption{Effect of the number of large tokens $n^{'}$. CSTA efficiently boost performance.}
    \label{efficientcsta}
    \end{subtable}
    \hfill
    \begin{subtable}[t]{0.28\textwidth}
    \centering
        \begin{adjustbox}{width=1\linewidth}
        \begin{tabular}[t]{llc}\toprule
        \textBF{\# Scales} & \textBF{Token sizes}
         & \textBF{Set14/Urban100} \\ \midrule
        1 & (3) & 34.45/33.93 \\ 
        2 & (3, 6) & \textBF{34.60}/\textBF{34.07} \\ 
        3 & (3, 6, 12) & 34.52/33.95 \\ \bottomrule
        \end{tabular}
        \end{adjustbox}
    \caption{Effect of various scaled tokens. CSTA with two scales performs best.}
    \label{largetoken}
    \end{subtable}
\caption{Ablation experiments on CSTA module. We demonstrate the effectiveness of our proposed CSTA on various aspects w.r.t. PSNR metric.}
\label{tab:csta}
\end{table*}

In \tabref{tab:design}, we ablate various architectural choices related to multi-stream network.
Specifically, we conduct experiments on each branch and fusion strategies.
First, due to the large model capacity, a single-stream network with only a transformer branch performs better than a relatively lightweight single-stream CNN branch.
Next, we observe that the performance of a two-stream network without Fusion Block consistently drops due to significantly separated pathways.
However, performance is largely improved when intermediate features are unidirectionally fused. 
Finally, the proposed bidirectional fusion between CNN and transformer features (T $\leftrightarrow$ C) shows the best performance over the entire fusion strategy.
The experimental results show that transformer and CNN branches contain complementary information, and intermediate bidirectional fusion is necessary for satisfactory restoration results.

\subsubsection{Impact of CSTA}
\label{csta}


\paragraph{MHSA vs. CSTA:} 

We compare our CSTA against standard MHSA in \tabref{cstamhsa}.
Specifically, we train ACT by replacing all CSTA/MHSA with MHSA/CSTA (\ie, \textit{MHSA only} and \textit{CSTA only}).
The comparison between \textit{MHSA only} and \textit{MHSA + CSTA} shows that $\text{CSTA}$ largely boosts performance with a small number of additional parameters (+ 0.7M).
Moreover, the result of \textit{CSTA only} indicates that CSTA alone cannot cover the role of MHSA capturing self-similarity within the same scale (performs better than \textit{MHSA only} but lower than \textit{MHSA + CSTA}).

\vspace{-0.2cm}
\paragraph{Impact of the number of large tokens:}

We conduct experiments to observe large tokens' impact and efficiency in \tabref{efficientcsta}.
Specifically, we vary sequence length of large token ($\ie, n^{\prime}$) of $\mathbf{T}^{l}$ by controlling stride $s^{\prime}$.
The results show that our CSTA module, even with a small number of large tokens ($n^{\prime}=81$), efficiently outperforms conventional self-attention without cross-attention (\ie, MHSA) by 0.06dB.
Furthermore, performance improvement is remarkable when we increase the number of large tokens with a small overhead in terms of FLOPs.
This experimental result demonstrates that CSTA can efficiently exploit informative cross-scale features with larger tokens.

\vspace{-0.2cm}
\paragraph{Impact of more token scales:}

We investigate whether performing CSTA with more token sizes is beneficial or not in \tabref{largetoken}.
In doing so, we embed three token sizes (3, 6, and 12) with similar overall computational costs to the CSTA module.
Comparing two token scales and three token scales shows that cross attention with an additional larger scale drops the performance.
Since the number of recurring patches decreases as scale increases~\cite{zontak2011internal}, we observe that exploiting self-similar patches across proper scales is more effective than solely performing certain types of attention or adding various scales.

\begin{table*}[t]
\scriptsize
\centering
\begin{adjustbox}{width=1\linewidth}
\begin{tabular}{lccccccccccc}\toprule
 &  & \multicolumn{2}{c}{\textBF{Set5} \cite{bevilacqua2012set5}} & \multicolumn{2}{c}{\textBF{Set14} \cite{zeyde2010set14}} & \multicolumn{2}{c}{\textBF{B100} \cite{martin2001b100}} & \multicolumn{2}{c}{\textBF{Urban100} \cite{huang2015selfex}} & \multicolumn{2}{c}{\textBF{Manga109} \cite{matsui2017manga109}} \\ \cmidrule(lr){3-4}\cmidrule(lr){5-6}\cmidrule(lr){7-8} \cmidrule(lr){9-10} \cmidrule(lr){11-12}  
\textBF{Method} & \textBF{Scale} & \textBF{PSNR} & \textBF{SSIM} & \textBF{PSNR} & \textBF{SSIM} & \textBF{PSNR} & \textBF{SSIM} & \textBF{PSNR} & \textBF{SSIM} & \textBF{PSNR} & \textBF{SSIM} \\ \midrule
EDSR \cite{lim2017edsr} & $\times 2$ & 38.11 & 0.9602 & 33.92 & 0.9195 & 32.32 & 0.9013 & 32.93 & 0.9351 & 39.10 & 0.9773 \\
RCAN \cite{zhang2018rcan} & $\times 2$ & 38.27 & 0.9614 & 34.12 & 0.9216 & 32.41 & 0.9027 & 33.34 & 0.9384 & 39.44 & 0.9786 \\
RNAN \cite{zhang2019rnan} & $\times 2$ & 38.17 & 0.9611 & 33.87 & 0.9207 & 32.32 & 0.9014 & 32.73 & 0.9340 & 39.23 & 0.9785 \\
SAN \cite{dai2019san} & $\times 2$ & 38.31 & 0.9620 & 34.07 & 0.9213 & 32.42 & 0.9028 & 33.10 & 0.9370 & 39.32 & 0.9792 \\
HAN \cite{niu2020han} & $\times 2$ & 38.27 & 0.9614 & 34.16 & 0.9217 & 32.41 & 0.9027 & 33.35 & 0.9385 & 39.46 & 0.9785 \\
NLSA \cite{mei2021nlsa} & $\times 2$ & 38.34 & 0.9618 & 34.08 & 0.9231 & 32.43 & 0.9027 & 33.42 & 0.9394 & 39.59 & 0.9789 \\ 
IPT \cite{chen2021ipt} & $\times 2$ & 38.37 & - & 34.43 & - & 32.48 & - & 33.76 & - & - & - \\
SwinIR \cite{liang2021swinir} & $\times 2$ & 38.42 & 0.9623 & 34.46 & 0.9250 & 32.53 & 0.9041 & 33.81 & 0.9427 & 39.92 & 0.9797 \\
ACT (Ours) & $\times 2$ & \underline{38.46} & \underline{0.9626} & \underline{34.60} & \underline{0.9256} & \underline{32.56} & \underline{0.9048} & \underline{34.07} & \underline{0.9443} & \underline{39.95} & \underline{0.9804} \\ 
ACT+ (Ours) & $\times 2$ & \textBF{38.53} & \textBF{0.9629} & \textBF{34.68} & \textBF{0.9260} & \textBF{32.60} & \textBF{0.9052} & \textBF{34.25} & \textBF{0.9453} & \textBF{40.11} & \textBF{0.9807} \\ \midrule
EDSR \cite{lim2017edsr} & $\times 3$& 34.65 & 0.9280 & 30.52 & 0.8462 & 29.25 & 0.8093 & 28.80 & 0.8653 & 34.17 & 0.9476 \\
RCAN \cite{zhang2018rcan} & $\times 3$& 34.74 & 0.9299 & 30.65 & 0.8482 & 29.32 & 0.8111 & 29.09 & 0.8702 & 34.44 & 0.9499 \\
SAN \cite{dai2019san} & $\times 3$& 34.75 & 0.9300 & 30.59 & 0.8476 & 29.33 & 0.8112 & 28.93 & 0.8671 & 34.30 & 0.9494 \\
HAN \cite{niu2020han} & $\times 3$& 34.75 & 0.9299 & 30.67 & 0.8483 & 29.32 & 0.8110 & 29.10 & 0.8705 & 34.48 & 0.9500 \\
NLSA \cite{mei2021nlsa} & $\times 3$& 34.85 & 0.9306 & 30.70 & 0.8485 & 29.34 & 0.8117 & 29.25 & 0.8726 & 34.57 & 0.9508 \\
IPT \cite{chen2021ipt} & $\times 3$ & 34.81 & - & 30.85 & - & 29.38 & - & 29.38 & - & - & - \\
SwinIR \cite{liang2021swinir} & $\times 3$ & 34.97 & 0.9318 & 30.93 & 0.8534 & 29.46 & 0.8145 & 29.75 & 0.8826 & 35.12 & 0.9537 \\
ACT (Ours) & $\times 3$ & \underline{35.03} & \underline{0.9321} & \underline{31.08} & \underline{0.8541} & \underline{29.51} & \underline{0.8164} & \underline{30.08} & \underline{0.8858} & \underline{35.27} & \underline{0.9540} \\
ACT+ (Ours) & $\times 3$ & \textBF{35.09} & \textBF{0.9325} & \textBF{31.17} & \textBF{0.8549} & \textBF{29.55} & \textBF{0.8171} & \textBF{30.26} & \textBF{0.8876} & \textBF{35.47} & \textBF{0.9548} \\ \midrule
EDSR \cite{lim2017edsr} & $\times 4$ & 32.46 & 0.8968 & 28.80 & 0.7876 & 27.71 & 0.7420 & 26.64 & 0.8033 & 31.02 & 0.9148 \\
RCAN \cite{zhang2018rcan} & $\times 4$ & 32.63 & 0.9002 & 28.87 & 0.7889 & 27.77 & 0.7436 & 26.82 & 0.8087 & 31.22 & 0.9173 \\
RNAN \cite{zhang2019rnan} & $\times 4$ & 32.49 & 0.8982 & 28.83 & 0.7878 & 27.72 & 0.7421 & 26.61 & 0.8023 & 31.09 & 0.9149 \\
SAN \cite{dai2019san}& $\times 4$ & 32.64 & 0.9003 & 28.92 & 0.7888 & 27.78 & 0.7436 & 26.79 & 0.8068 & 31.18 & 0.9169 \\
HAN \cite{niu2020han} & $\times 4$ & 32.64 & 0.9002 & 28.90 & 0.7890 & 27.80 & 0.7442 & 26.85 & 0.8094 & 31.42 & 0.9177 \\
NLSA \cite{mei2021nlsa} & $\times 4$ & 32.59 & 0.9000 & 28.87 & 0.7891 & 27.78 & 0.7444 & 26.96 & 0.8109 & 31.27 & 0.9184 \\
IPT \cite{chen2021ipt} & $\times 4$ & 32.64 & - & 29.01 & - & 27.82 & - & 27.26 & - & - & - \\
SwinIR \cite{liang2021swinir} & $\times 4$ & 32.92 & \textBF{0.9044} & 29.09 & 0.7950 & 27.92 & 0.7489 & 27.45 & 0.8254 & 32.03 & 0.9260 \\
ACT (Ours) & $\times 4$ & \underline{32.97} & 0.9031 & \underline{29.18} & \underline{0.7954} & \underline{27.95} & \underline{0.7507} & \underline{27.74} & \underline{0.8305} & \underline{32.20} & \underline{0.9267} \\ 
ACT+ (Ours) & $\times 4$ & \textBF{33.04} & \underline{0.9041} & \textBF{29.27} & \textBF{0.7968} & \textBF{28.00} & \textBF{0.7516} & \textBF{27.92} & \textBF{0.8332} & \textBF{32.44} & \textBF{0.9282} \\ \bottomrule
\end{tabular}
\end{adjustbox}
\caption{Quantitative comparison of the proposed method with numerous state-of-the-art SR methods. The best and the second-best values are highlighted with \textBF{bold} and \underline{underline}, respectively.}
\label{tab:main}
\end{table*}
  
\subsubsection{Feature visualization}
\label{featvis}

\begin{figure}[t]
    \centering
    \includegraphics[width=\linewidth]{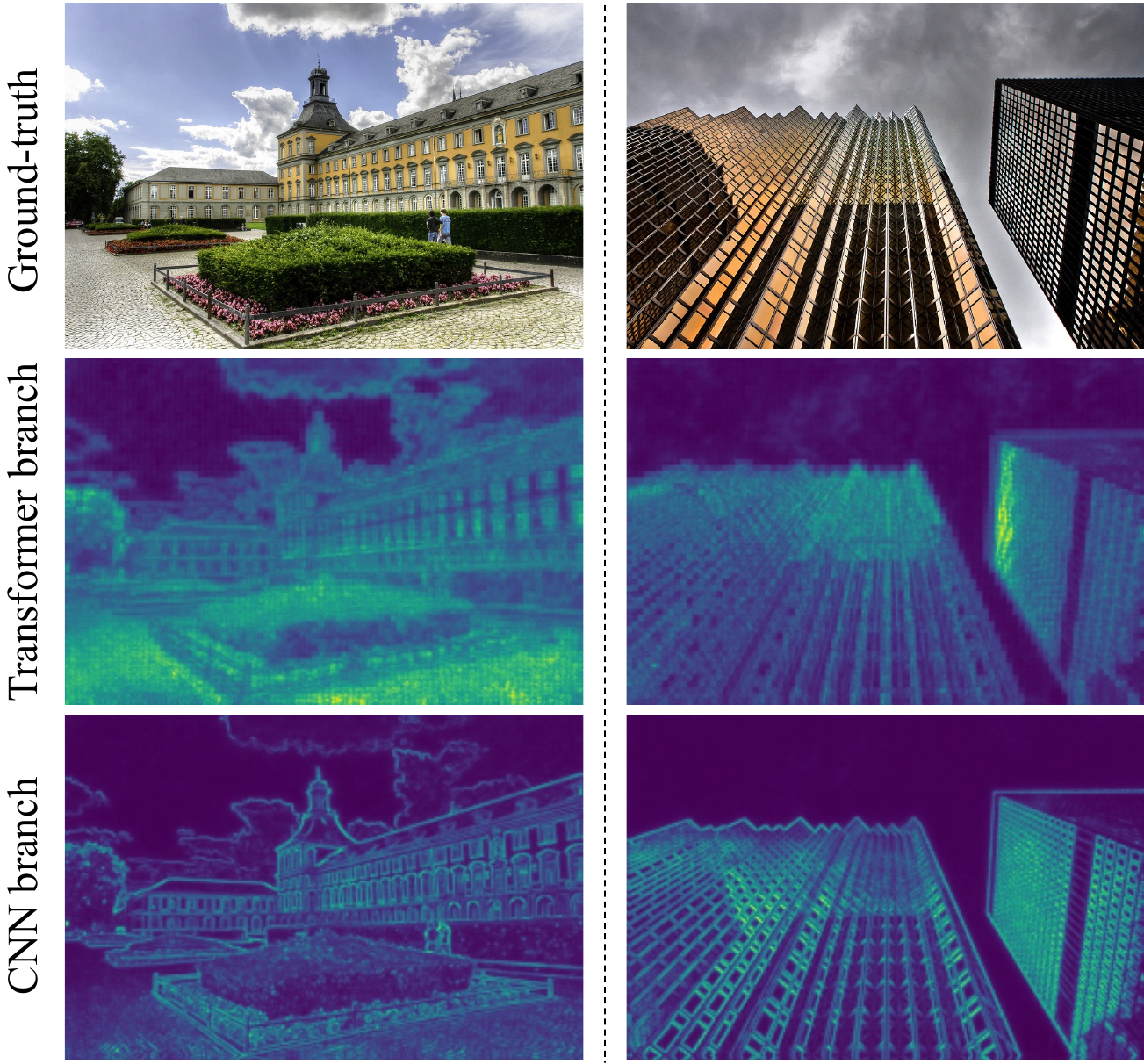}
    \caption{Feature map visualizations of transformer branch and CNN branch. The transformer branch focuses on restoring texture detail and repeated small patterns, while the CNN branch emphasizes the reconstruction of sharp and strong edges.}
    \label{fig:feature}
\end{figure}

In \figref{fig:feature}, we visualize features to analyze the role of each branch.
Specifically, we compare the output features from the last blocks for each branch (\ie, $\textbf{T}_{4}$ and $\textbf{F}_{4}$).
According to the visualization, both branches provide minimal attention to flat and low-frequency areas (\eg, sky).
However, the output feature from the transformer branch focuses on recovering tiny and high-frequency texture details, while producing blurry and checkerboard artifacts due to tokenization.
Moreover, we observe that the transformer branch attends to a small version of recurring patches within the image (\eg, upper side window in the right example), leveraging multi-scale representations with $\text{CSTA}$ module.
In contrast, the CNN branch recovers sharp and strong edges, which the transformer branch fails to capture.
This observation indicates that ACT endowed independent roles for each pathway such that the two branches complement each other.

\subsection{SR results}

\begin{figure*}[t]
    \centering
    \includegraphics[width=1\linewidth]{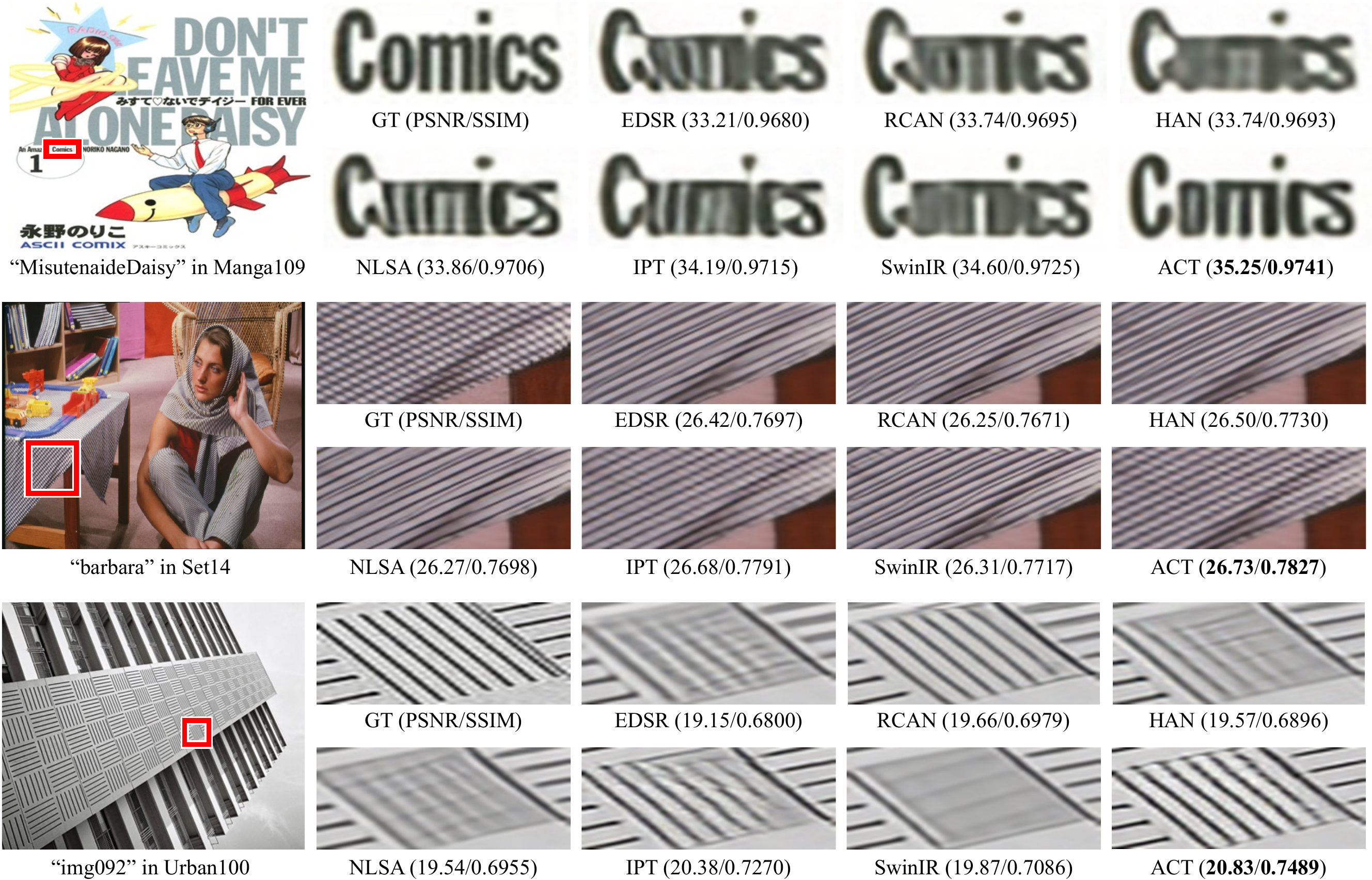}
    \caption{Visual comparison of the proposed method with various methods for $\times 4$ SR task. Our method restores sharp and complicated structures more accurately.}
    \label{fig:visualize}
\end{figure*}

\paragraph{Quantitative evaluation:}

In \tabref{tab:main}, we quantitatively compare our ACT for $\times 2$, $\times 3$, and $\times 4$ SR tasks with eight state-of-the-art SR networks: EDSR \cite{lim2017edsr}, RCAN \cite{zhang2018rcan}, RNAN \cite{zhang2019rnan}, SAN \cite{dai2019san}, HAN \cite{niu2020han}, NLSA \cite{mei2021nlsa}, IPT \cite{chen2021ipt}, and SwinIR \cite{liang2021swinir}.
We also report test-time self-ensembling-based results following baselines \cite{lim2017edsr, zhang2018rcan, niu2020han, liang2021swinir} to improve performance further, and ACT+ indicates approach with self-ensemble \cite{lim2017edsr}.
Compared with all previous works, ACT and ACT+ achieve the best or second-best performance in terms of PSNR and SSIM for all scale factors.
In particular, our method substantially outperforms IPT \cite{chen2021ipt}, which is recognized as the first transformer-based restoration approach, 
through proposed multi-scale feature extraction and effective hybridized architecture of CNN and transformer.
Moreover, performance improvement over SwinIR \cite{liang2021swinir} is considerable for Urban100 dataset \cite{huang2015selfex} (more than 0.3dB PSNR gain for all scale factors) with high patch-recurrence in the dataset, indicating that our CSTA module successfully exploits multi-scale features.

\vspace{-0.2cm}
\paragraph{Qualitative evaluation:}

We provide a qualitative comparison with existing SR methods.
 \figref{fig:visualize} shows that our method obtains more accurately recovered details than conventional methods.
Specifically, the restoration result of image ``MisutenaideDaisy" demonstrates that our method can generate more human-readable characters than other existing methods.
Moreover, by taking ``barbara" as an example, baseline methods have generated sharp edges/patterns despite being far from the ground-truth structure. 
By contrast, our method correctly reconstructs the main structure without losing high-frequency details.
The result of ``img092", which contains an urban scene, shows that most conventional methods fail to recover the structure and produce blurry results.
Meanwhile, our method alleviates blurring artifacts and accurately reconstructs correct contents.
The above observation indicates the general superiority of the proposed method in recovering sharp and accurate details.

\subsection{Model size analysis}


Finally, we compare the number of network parameters and floating-point operations (FLOPs) of various SR methods in~\tabref{tab:model_size}.
Our ACT shows the best SR results as in~\tabref{tab:main} with competitive hardware resources in comparison with existing approaches, including IPT~\cite{chen2021ipt} and SwinIR~\cite{liang2021swinir}.
Notably, although SwinIR~\cite{liang2021swinir} has few parameters, its computational cost is relatively high due to small window and token sizes, 8$\times$8 and 1$\times$1, respectively.
The comparison demonstrates an effective trade-off between ACT's performance and model complexity.

\section{Conclusion}

\begin{table}[t]
    \centering
    \begin{adjustbox}{width=\linewidth}
    \begin{tabular}{lcccccc}\toprule
    \textBF{Measure} & \textBF{EDSR} & \textBF{RCAN} & \textBF{NLSA} & \textBF{IPT} & \textBF{SwinIR} & \textBF{ACT} (Ours) \\ \midrule
    \# Params. & 43M & 16M & 44M & 114M & 12M & 46M \\
    FLOPs & 116G & 37G & 125G & 35G & 29G & 22G \\ \bottomrule
    \end{tabular}
    \end{adjustbox}
\caption{Comparison of the proposed method's resources with state-of-the-art SR methods.}
\vspace{-0.2cm}
\label{tab:model_size}
\end{table}

In this study, we proposed to aggregate various beneficial features for SR and introduced a novel architecture combining transformer and convolutional branches, advantageously fusing both representations.
Moreover, we presented an efficient cross-scale attention module to exploit multi-scale feature maps within a transformer branch.
The effectiveness of the proposed method has been extensively demonstrated under numerous benchmark SR datasets, and our method records the state-of-the-art SR performance in terms of quantitative and qualitative comparisons.

\subsubsection*{Acknowledgments}
This work was partially supported by Institute of Information \& communications Technology Planning \& Evaluation (IITP) grant funded by the Korea government(MSIT) (No.2022-0-00156, Fundamental research on continual meta-learning for quality enhancement of casual videos and their 3D metaverse transformation.)


{\small
\bibliographystyle{ieee_fullname}
\bibliography{egbib}
}

\section*{Appendix}

\appendix

In this appendix, we provide additional details, experimental settings, and results of our proposed method.
First, we describe the details of the proposed Fusion Block and CSTA.
Next, we report more results related to the model size analysis.
Then, we show the applicability of our approach to another image restoration task (\ie, JPEG artifact removal).
Finally, we provide more visual results of feature visualization and qualitative comparisons.

\section{Fusion Block}
\label{sec:fb}

\paragraph{Settings for the ablation:}

\begin{figure*}[t]
    \centering
    \includegraphics[width=0.8\linewidth]{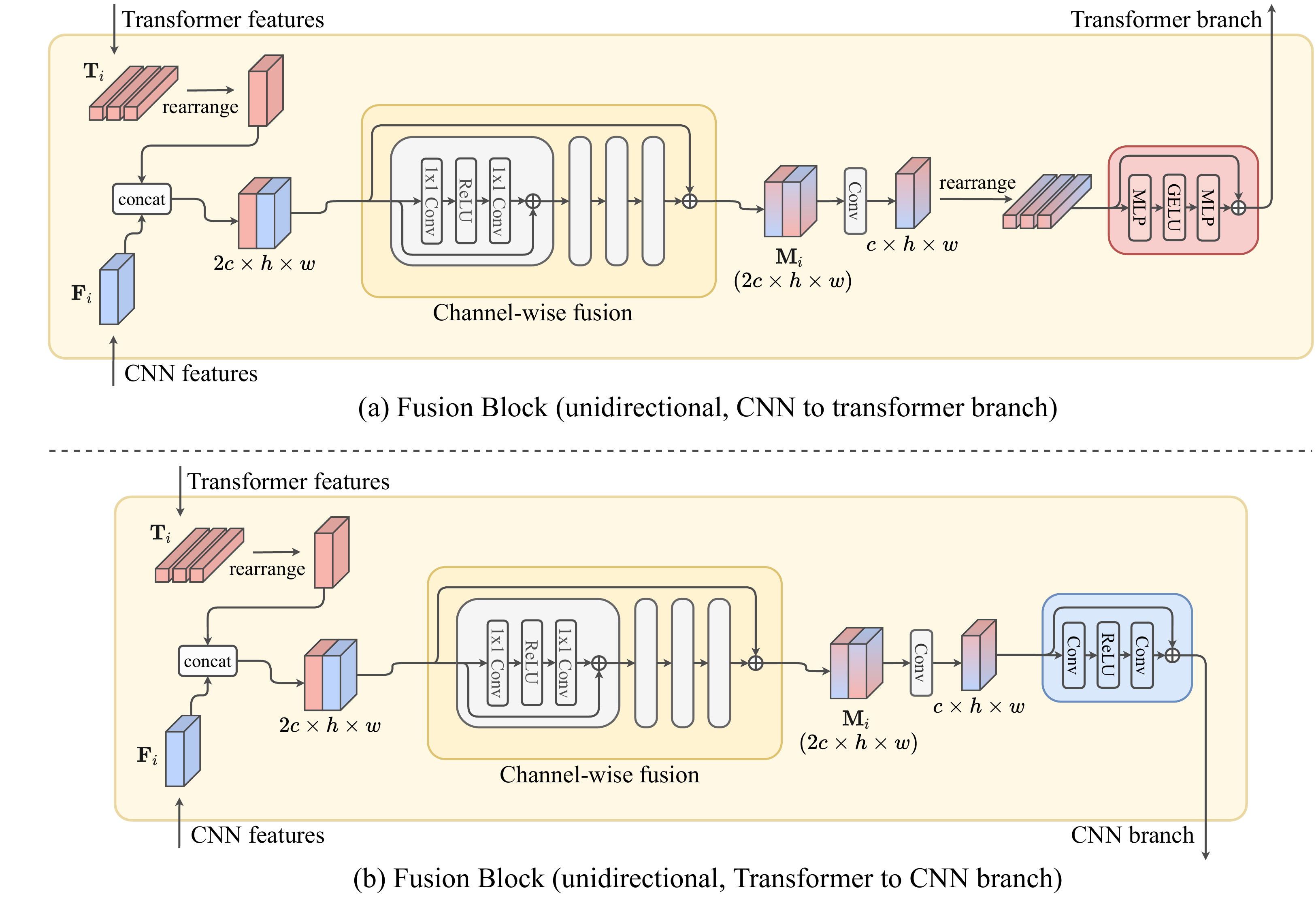}
    \vspace{-0.2cm}
    \caption{Illustration of unidirectional Fusion Blocks, used for the main manuscript's ablation experiments.}
    \label{fig:uni}
\end{figure*}

We elaborate on detailed settings of our fusion strategies.
For the ablation experiment without intermediate fusion (third column in \tabref{tab:design}), we concatenate output features $\mathbf{T}_4$ and $\mathbf{F}_4$ from separated branches and pass them to the tail module.
For the experiments with the unidirectional flow (fourth and fifth columns in \tabref{tab:design}), we do not split feature $\mathbf{M}_i$ into two.
Instead, we reduce channel dimension from $2c$ to $c$ with a single convolution layer and transfer it to branch chosen to receive fused representation (see \figref{fig:uni}).

\paragraph{Experiment on lateral connection:}

Since the dimension of the feature from the CNN branch and image-likely rearranged feature from the transformer branch are the same (\ie, $c \times h \times w$), one can laterally connect two features with element-wise summation rather than concatenation in Equation \eqref{eq:7}.
We compare the performance of different lateral connection choices in \tabref{tab:sum} and observe that concatenation gives slightly better results than summation.
Therefore, we finalize our lateral connection to concatenating two features.

\section{Cross-Scale Token Attention}
\label{sec:csta}

\paragraph{Pseudocode of CSTA:}

\algoref{algo:csta} provides pseudocode of CSTA for better understanding.
In particular, we split the hidden dimension of $\mathbf{T}$ into two and directly utilize $\mathbf{T}^a$ for $\mathbf{T}^s$ while reformulating $\mathbf{T}^b$ to acquire $\mathbf{T}^l$ as elaborated in the main manuscript.
It is worth noting that such rearrangement is similar to soft-split, which is introduced in T2T-ViT \cite{yuan2021t2t} in that tokens are re-structured by overlapping.
Unlike soft-split, we rearrange only a part of the hidden dimension within token $\mathbf{T}$ with a larger token size to perform cross-attention efficiently. 
Moreover, our rearrangement aims to acquire numerous larger tokens (patches) to exploit recurring patches across different scales within the input image.

\paragraph{Settings for the ablation:}

We elaborate details about \textit{Impact of more token scales} experiment in our ablation on CSTA (\tabref{largetoken}).
Specifically, we demonstrate how we perform cross-attention by introducing three different token scales (3rd row in \tabref{largetoken}).
To enable our network to leverage patch-recurrence across various scales while maintaining similar computational cost, we first split input token $\mathbf{T} \in \mathbb{R}^{n \times d}$ into four tokens $\{\mathbf{T}_i\}_{i=1}^{4} \in \mathbb{R}^{n \times \sfrac{d}{4}}$ before cross-attention.
Then, we rearrange $\mathbf{T}_{2}$ and $\mathbf{T}_{4}$ with token size of $6 \times 6$ and $12 \times 12$, respectively, with the same strides of 3, while $\mathbf{T}_{1}$ and $\mathbf{T}_{3}$ keep their token size of $3\times3$.
We perform two independent cross-attention using ($\mathbf{T}_{1}$, $\mathbf{T}_{2}$) and ($\mathbf{T}_{3}$, $\mathbf{T}_{4}$) pair as elaborated in the main manuscript.
Consequently, the network can utilize multi-scale information across various scales.
Lastly, we re-concatenate four tokens into one after rearranging token $\mathbf{T}_{2}$ and $\mathbf{T}_{4}$ to include a token size of $3\times3$.

\begin{table}[t]
\centering
    \begin{adjustbox}{width=0.95\linewidth}
    \begin{tabular}{lcc} \\ \toprule
    \textBF{Method} & \textBF{Lateral connection} & \textBF{Set14/Urban100} \\ \midrule
    ACT (Ours) & element-wise summation & 34.57/34.05 \\
    ACT (Ours) & concatenation & \textBF{34.60}/\textBF{34.07} \\ \bottomrule
    \end{tabular}
    \end{adjustbox}
\caption{Comparison on different lateral connections of two branches in our Fusion Block, reported in PSNR value.}
\label{tab:sum}
\end{table}

\section{Model Size Analysis}
\label{sec:model}

\begin{table}
\centering
\begin{adjustbox}{width=0.6\linewidth}
\begin{tabular}{ccc} \toprule
\textBF{IPT} \cite{chen2021ipt} & \textBF{SwinIR} \cite{liang2021swinir} & \textBF{ACT} (Ours) \\ \midrule
0.586s & 0.528s & 0.566s \\ \bottomrule
\end{tabular}
\end{adjustbox}
\caption{Runtime comparison.}
\label{runtime}
\end{table}

For model size analysis in \tabref{tab:model_size}, we count FLOPs for $48 \times 48$ image following IPT \cite{chen2021ipt} using the open source. \footnote{\href{https://github.com/facebookresearch/fvcore}{https://github.com/facebookresearch/fvcore}}
Moreover, we compare the runtime of our model with recently proposed transformer-based SR methods \cite{chen2021ipt, liang2021swinir}.
We average five runs on the same setting as in \tabref{tab:model_size} with Ryzen 2950X CPU and NVIDIA 2080 Ti.
The result in \tabref{runtime} shows that ours is competitive in runtime.

\section{Applicability to other Image Restoration Tasks}
\label{sec:car}

\begin{table*}[h]
\centering
\begin{adjustbox}{width=0.8\linewidth}
\begin{tabular}{lx{24}|x{80}x{80}x{80}x{80}} \toprule
\textBF{Dataset} & \textBF{QF} & \textBF{JPEG} & \textBF{QGAC} \cite{ehrlich2020qgac} & \textBF{FBCNN-C} \cite{jiang2021fbcnn} & \textBF{ACT} (Ours) \\ \midrule
\multirow{4}{*}{LIVE1 \cite{sheikh2005live}} & 10 & 25.69/0.743/24.20 & 27.62/\underline{0.804}/27.43 & \underline{27.77}/0.803/\underline{27.51} & \textBF{27.94}/\textBF{0.808}/\textBF{27.63} \\
 & 20 & 28.06/0.826/26.49 & 29.88/\underline{0.868}/29.56 & \underline{30.11}/\underline{0.868}/\underline{29.70} & \textBF{30.39}/\textBF{0.874}/\textBF{29.96} \\
 & 30 & 29.37/0.861/27.84 & 31.17/0.896/30.77 & \underline{31.43}/\underline{0.897}/\underline{30.92} & \textBF{31.77}/\textBF{0.902}/\textBF{31.26} \\
 & 40 & 30.28/0.882/28.84 & 32.05/0.912/31.61 & \underline{32.34}/\underline{0.913}/\underline{31.80} & \textBF{32.70}/\textBF{0.917}/\textBF{32.16} \\ \midrule
\multirow{4}{*}{BSDS500 \cite{martin2001bsds500}} & 10 & 25.84/0.741/24.13 & 27.74/\textBF{0.802}/\underline{27.47} & \textBF{27.85}/0.799/\textBF{27.52} & \underline{27.84}/\underline{0.800}/27.46 \\
 & 20 & 28.21/0.827/26.37 & 30.01/\textBF{0.869}/29.53 & \underline{30.14}/0.867/\underline{29.56} & \textBF{30.20}/\underline{0.868}/\textBF{29.58} \\
 & 30 & 29.57/0.865/27.72 & 31.33/\textBF{0.898}/30.70 & \underline{31.45}/\underline{0.897}/\underline{30.72} & \textBF{31.51}/\underline{0.897}/\textBF{30.77} \\
 & 40 & 30.52/0.887/28.69 & 32.25/\textBF{0.915}/31.50 & \underline{32.36}/0.913/\underline{31.52} & \textBF{32.41}/\underline{0.914}/\textBF{31.56} \\ \midrule
\multirow{4}{*}{ICB \cite{icb}} & 10 & 29.44/0.757/28.53 & 32.06/\textBF{0.816}/32.04 & \underline{32.18}/\underline{0.815}/\underline{32.15} & \textBF{32.20}/\textBF{0.816}/\textBF{32.17} \\
 & 20 & 32.01/0.806/31.11 & 34.13/\underline{0.843}/34.10 & \underline{34.38}/\textBF{0.844}/\underline{34.34} & \textBF{34.50}/\textBF{0.844}/\textBF{34.46} \\
 & 30 & 33.20/0.831/32.35 & 35.07/\underline{0.857}/35.02 & \underline{35.41}/\underline{0.857}/\underline{35.35} & \textBF{35.61}/\textBF{0.859}/\textBF{35.55} \\
 & 40 & 33.95/0.840/33.14 & 32.25/\textBF{0.915}/31.50 & \underline{36.02}/0.866/\underline{35.95} & \textBF{36.21}/\underline{0.868}/\textBF{36.13} \\ \bottomrule
\end{tabular}
\end{adjustbox}
\caption{PSNR/SSIM/PSNRB comparison of different state-of-the-art methods on color JPEG artifact removal. Our ACT shows competitive performance over baselines. Performances for the baselines are borrowed from \cite{jiang2021fbcnn}. The best and the second-best values are highlighted with \textBF{bold} and \underline{underline}, respectively.}
\label{tab:car}
\end{table*}

\begin{figure*}[t]
    \centering
    \includegraphics[width=1\linewidth]{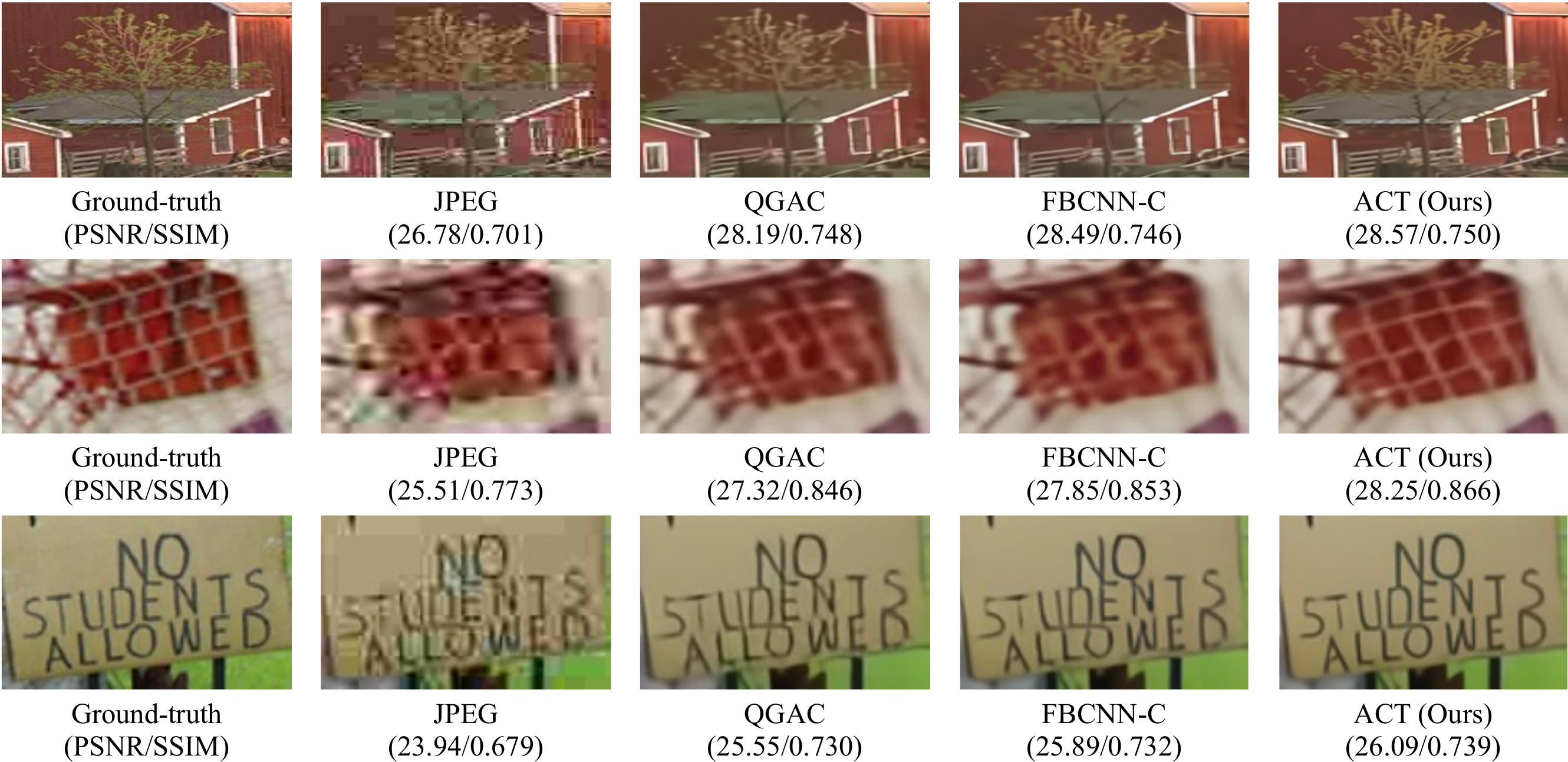}
    \caption{Visual comparison of our ACT against state-of-the-art color JPEG artifact removal methods \cite{ehrlich2020qgac, jiang2021fbcnn} with quality factor of 10. Our ACT better removes the artifacts and produces accurate structures than the baselines.}
    \label{fig:car}
\end{figure*}

We investigate the applicability of our network to a challenging restoration task; color JPEG artifact removal.
To do so, we only modify the tail part of the network to include a single convolutional layer while discarding PixelShuffle upsampler~\cite{shi2016pixelshuffle}.
We train our network with the same training configurations for SR, and the training patch size is $48 \times 48$.
We compare our ACT with the state-of-the-art color JPEG artifact removal methods, including QGAC~\cite{ehrlich2020qgac} and FBCNN-C~\cite{jiang2021fbcnn} for quality factors of 10, 20, 30, and 40.
We evaluate performance on LIVE1~\cite{sheikh2005live}, BSDS500~\cite{martin2001bsds500}, and
ICB~\cite{icb} datasets with three different metrics including PSNR, SSIM, and PSNR-B values following the baselines~\cite{ehrlich2020qgac, jiang2021fbcnn}.
We report the quantitative comparison in \tabref{tab:car}.
As shown in the table, our ACT shows promising results despite lacking task-specific architectural design.
Moreover, we visually compare our method against the baselines in \figref{fig:car}.
Compared to the baselines, our ACT accurately restores corrupted images, including natural scenes, sharp textures, and characters.
This result implies the possibility of applying our ACT to various restoration tasks similar to the recent image restoration transformers~\cite{chen2021ipt, liang2021swinir, wang2021uformer}.

\begin{algorithm*}[h]
\SetAlgoLined
    \PyComment{$b$, $n$, $d$: batch size, number of tokens, and hidden dimension of $\mathbf{T}$} \\
    \PyComment{$n^{\prime}$, $d^{\prime}$: number of tokens and hidden dimension of $\mathbf{T}^{l}$ directly after acquiring large tokens from $\mathbf{T}^{b}$ by rearrangement} \\
    \PyComment{$h$, $w$: height and width of patches} \\
    \PyCode{} \\

    \PyCode{import torch} \\
    \PyCode{import torch.nn.functional as F} \\
    \PyCode{} \\

    \PyCode{def CSTA(T):} \\
    \Indp
        \PyComment{split $\mathbf{T}$ ($b \times n \times d$) into $\mathbf{T}^a$ and $\mathbf{T}^b$} \\
        \PyCode{T\_a, T\_b = torch.split(T, d//2, dim=2)} \\
        \PyComment{acquire $\mathbf{T}^s$ from $\mathbf{T}^a$} \\
        \PyCode{T\_s = T\_a} \\
        \PyComment{acquire $\mathbf{T}^l$ from $\mathbf{T}^b$ by rearrangement} \\
        \PyCode{T\_l = F.fold(T\_b, output\_size=(h, w), kernel\_size=token\_size, stride=token\_size)} \\
        \PyCode{T\_l = F.unfold(T\_l, kernel\_size=token\_size*2, stride=token\_size)} \\
        \PyCode{} \\
        
        \PyComment{project tokens into query, key, and value} \\
        \PyCode{T\_l = mlp\_blk\_before\_attn(T\_l)} \PyComment{reduce dimension from $d^{\prime}$ to $d/2$} \\
        \PyCode{q\_l, k\_l, v\_l = project(T\_l)} \\
        \PyCode{q\_s, k\_s, v\_s = project(T\_s)} \\
        \PyCode{} \\
    
        \PyComment{perform cross attention} \\
        \PyCode{T\_l = attention(query=q\_l, key=k\_s, value=v\_s)} \\
        \PyCode{T\_s = attention(query=q\_s, key=k\_l, value=v\_l)} \\
        \PyCode{} \\
        
        \PyComment{increase hidden dimension of $\mathbf{T}^l$ from $d/2$ to $d^{\prime}$} \\
        \PyCode{T\_l = mlp\_blk\_after\_attn(T\_l)} \\
        \PyCode{} \\
        
        \PyComment{rearrange $\mathbf{T}^l$ from ($b \times n^{\prime} \times d^{\prime}$) to ($b \times n \times (d/2)$)} \\
        \PyCode{T\_l = F.fold(T\_l, output\_size=(h, w), kernel\_size=token\_size*2, stride=token\_size)} \\
    
        \PyCode{T\_l = F.unfold(T\_l, kernel\_size=token\_size, stride=token\_size)} \\
        \PyCode{} \\
        
        \PyComment{concatenate $\mathbf{T}^s$ and $\mathbf{T}^l$ into $\mathbf{T}$ ($b \times n \times d$)} \\
        \PyCode{T = torch.cat((T\_s, T\_l), dim=2)} \\
        \PyCode{} \\
        
        \PyCode{return T}
        
    \Indm
\caption{Pseudocode of CSTA in a PyTorch-like style.}
\label{algo:csta}
\end{algorithm*}

\section{Feature Visualization}
\label{feature}

In \figref{fig:feature}, we provide more feature visualizations to understand the role of two branches.

\section{Additional Qualitative Results}
\label{qualitative}

\begin{figure*}[t]
    \centering
    \includegraphics[width=1\linewidth]{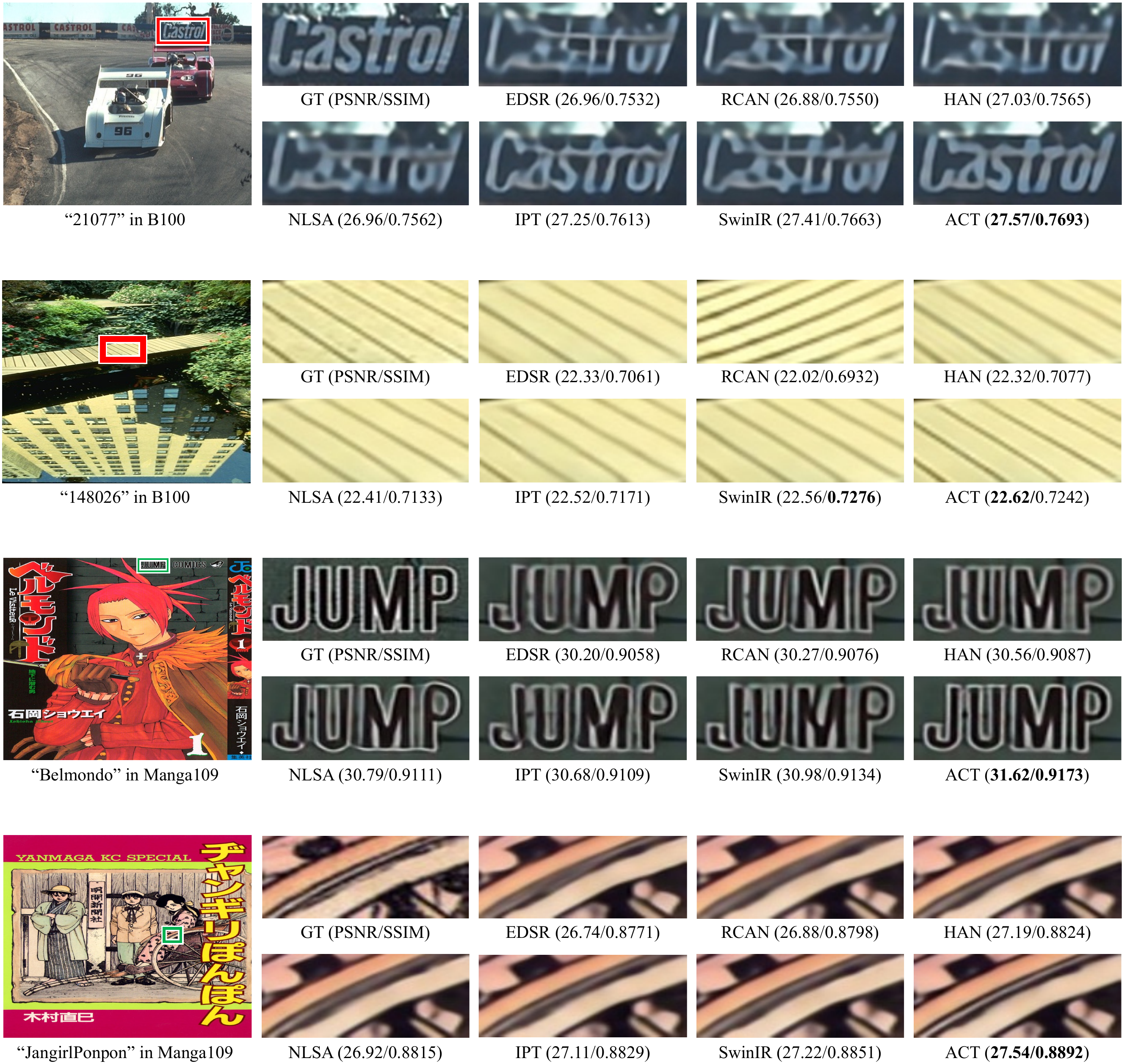}
    \caption{Visual comparison of the proposed method against various state-of-the-art methods for $\times 4$ SR.} 
    \label{fig:visualize1}
\end{figure*}

\begin{figure*}[t]
    \centering
    \includegraphics[width=1\linewidth]{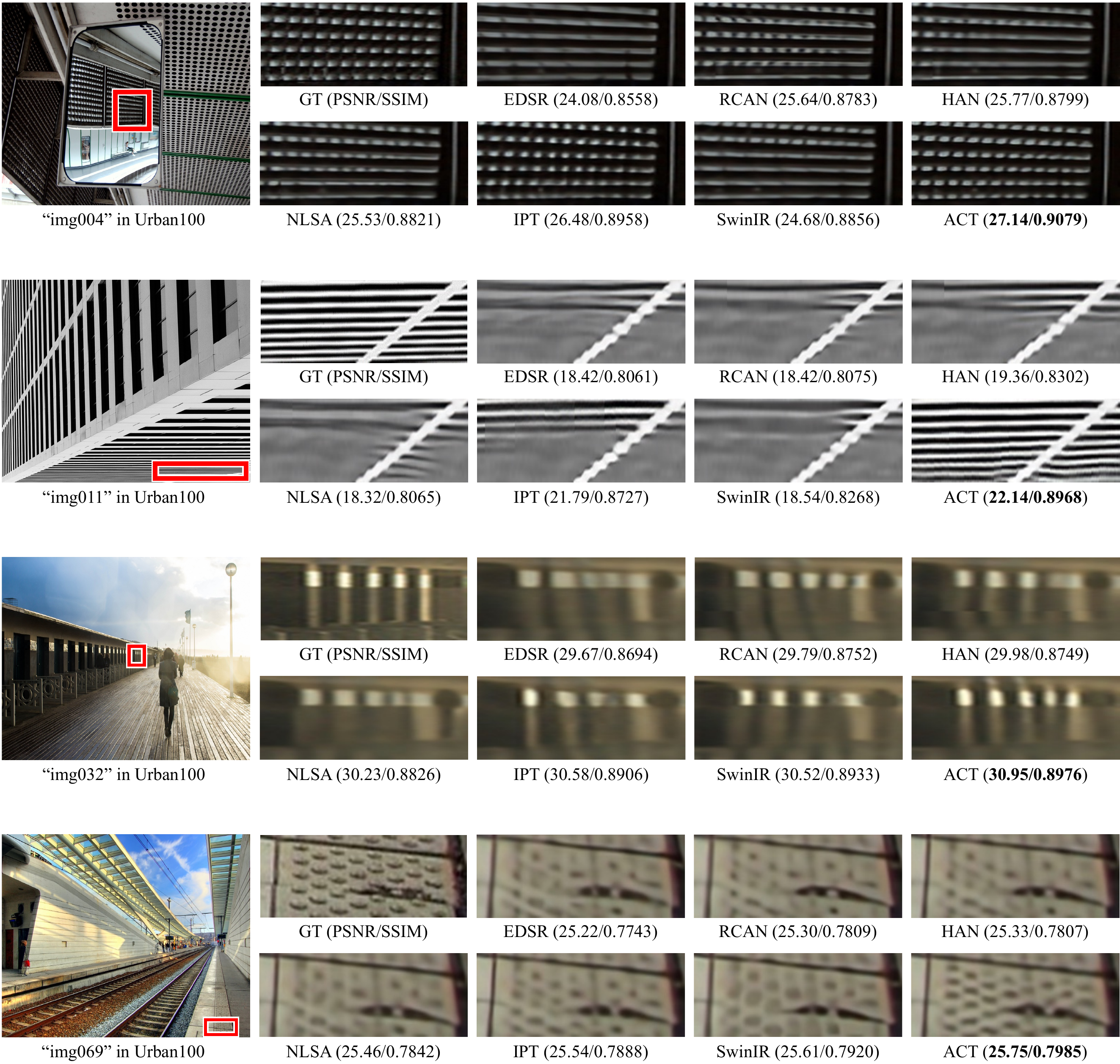}
    \caption{Visual comparison of the proposed method against various state-of-the-art methods for $\times 4$ SR.} 
    \label{fig:visualize2}
\end{figure*}

To further demonstrate the superiority of our proposed method,
we provide more visual comparisons with six state-of-the-art SR methods: EDSR~\cite{lim2017edsr}, RCAN~\cite{zhang2018rcan}, HAN~\cite{niu2020han}, NLSA~\cite{mei2021nlsa}, IPT~\cite{chen2021ipt}, and SwinIR~\cite{liang2021swinir}. 
The visual comparisons are shown in \figref{fig:visualize1} and \figref{fig:visualize2}.

\begin{figure*}[t]
    \centering
    \includegraphics[width=1\linewidth]{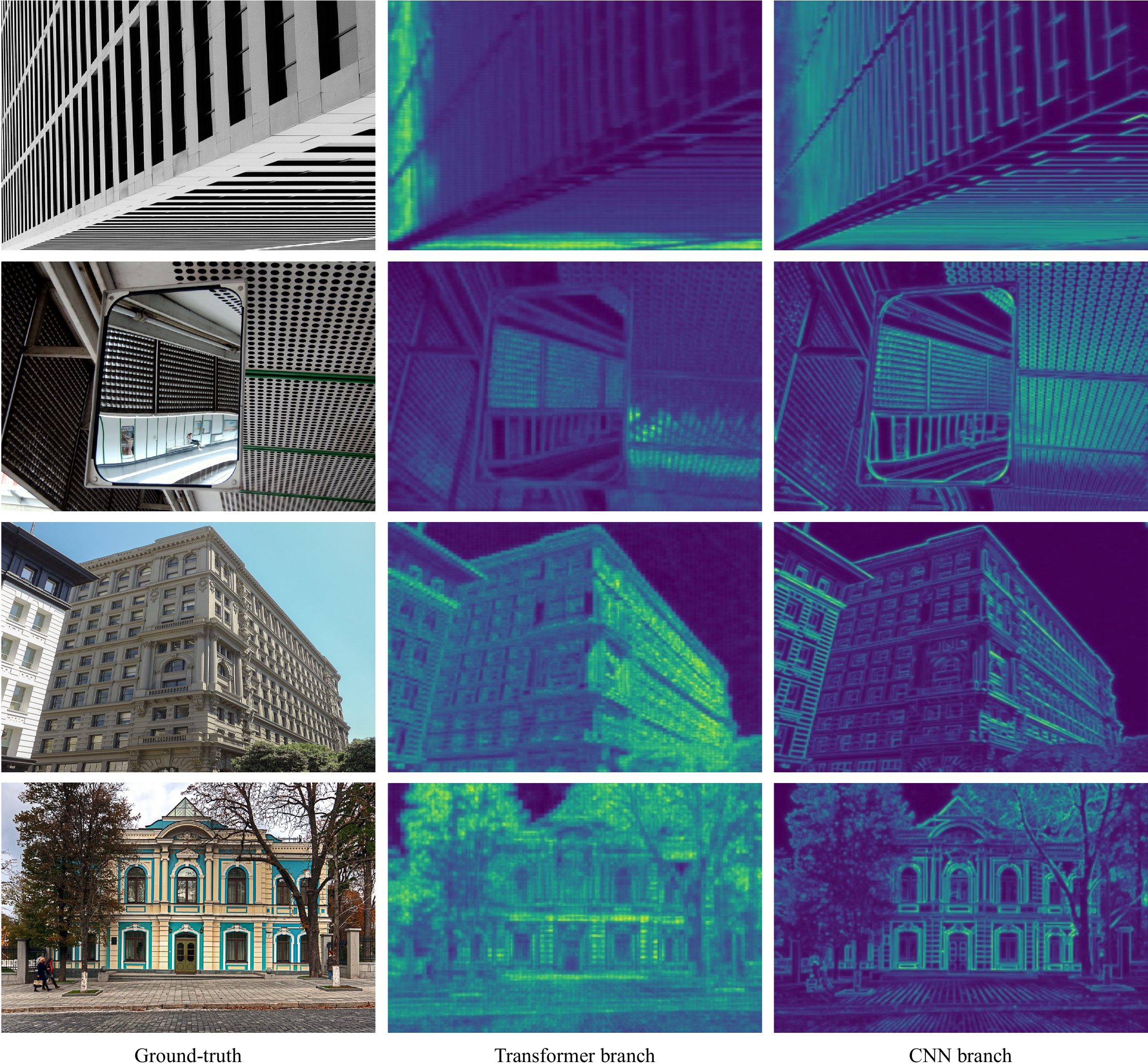}
    \caption{Feature map visualizations of transformer branch and CNN branch. Brighter color indicates higher value.}
    \label{fig:feature}
\end{figure*}

\end{document}